\documentclass{article}

\usepackage[final]{corl_2026} 
\usepackage{algorithm}
\usepackage{algpseudocode}
\usepackage{amsmath}
\usepackage{amssymb}
\usepackage{booktabs}
\usepackage{graphicx}
\usepackage{mathtools}
\usepackage{xcolor}
\usepackage{xparse}
\usepackage{tabularray}
\usepackage{tabularx}
\usepackage{wrapfig}

\def\papermode{draft}
\def\draftmode{draft}

\def\Snospace~{\S{}}

\definecolor[named]{TODORed}{HTML}{C62828}
\definecolor[named]{TLDRViolet}{HTML}{7B1FA2}
\definecolor[named]{MissingCyan}{HTML}{00838F}
\definecolor[named]{PendingOrange}{HTML}{EF6C00}
\definecolor[named]{cBlue}{HTML}{1565C0}
\definecolor[named]{cOrange}{HTML}{EF6C00}
\definecolor[named]{cPink}{HTML}{AD1457}
\definecolor[named]{cYellow}{HTML}{AF8F00}
\definecolor[named]{cGreen}{HTML}{2E7D32}
\definecolor[named]{cGray}{HTML}{616161}

\ifx\papermode\draftmode
    \newcommand{\todo}[1]{{\color{TODORed}#1}}
    \newcommand{\TODO}[1]{{\color{TODORed}\textbf{[TODO]} #1}}
    \newcommand{\tldr}[1]{\medskip\noindent{\color{TLDRViolet}\textbf{[TL;DR]} #1}}
    
    \newcommand{\placeholder}[1]{{\color{PendingOrange}\textbf{[placeholder]} #1}}
    \NewDocumentCommand{\makecomment}{m m m o}{%
      \noindent{\color{#2}[\textbf{#1}] #3}%
      \IfValueT{#4}{{\color{cGray}\space\textit{#4}}}%
    }
\else
    \newcommand{\todo}[1]{}
    \newcommand{\TODO}[1]{}
    \newcommand{\tldr}[1]{}
    
    \newcommand{\placeholder}[1]{}
    \NewDocumentCommand{\makecomment}{m m m o}{}
\fi

\NewDocumentCommand{\zhenyang}{m o}{\makecomment{ZC}{cYellow}{#1}[#2]}

\NewDocumentCommand{\woochul}{m o}{\makecomment{WS}{cPink}{#1}[#2]}

\NewDocumentCommand{\danfei}{m o}{\makecomment{DX}{cBlue}{#1}[#2]}

\usepackage{caption}
\usepackage{booktabs}
\title{Static In, Dynamic Out: Counterfactual Action Augmentation for Moving Object Manipulation}

\author{
  Woo Chul Shin,
  Zhenyang Chen,
  Alfred Cueva$^{*}$,
  Nadun Ranawaka$^{*}$, \\
  \textbf{Yingyan Celine Lin},
  \textbf{Benjamin Joffe},
  \textbf{Shreyas Kousik},
  \textbf{Danfei Xu} \\
  Georgia Institute of Technology
}

\begin{document}
\maketitle


\begin{abstract}
\label{sec:abstract}

Visuomotor policies have advanced on manipulation tasks where the target
object stays static during execution, but real deployments break this
assumption: parts drift on conveyors and fruits sway in the wind. We introduce \textbf{\textsc{Static In, Dynamic Out (SIDO)}}, a counterfactual action augmentation that enables a policy trained only on static object demonstrations to adapt to unseen object motion at test time. Our key idea is to factorize moving object manipulation into two sub-problems: predicting where the object will be, and reaching that predicted pose. SIDO displaces the object to a counterfactual future position and morphs the demonstrated action chunk to preserve the hand-object relative pose, yielding a goal-conditioned policy. At deployment an object pose predictor supplies the future position.  Across three simulated tasks (Mug, Square, Stack) under five object motion patterns and two real-world tasks (Gantry, Peachtree), SIDO improves moving object success over the baselines while preserving
static object performance. Project website: \href{https://sido-staticindynamicout.github.io/}{https://sido-staticindynamicout.github.io/}.

\end{abstract}



\section{Introduction}
\label{sec:introduction}

\vspace{-15pt}

\begin{figure}[ht]
    \centering
    \includegraphics[width=0.95\textwidth]{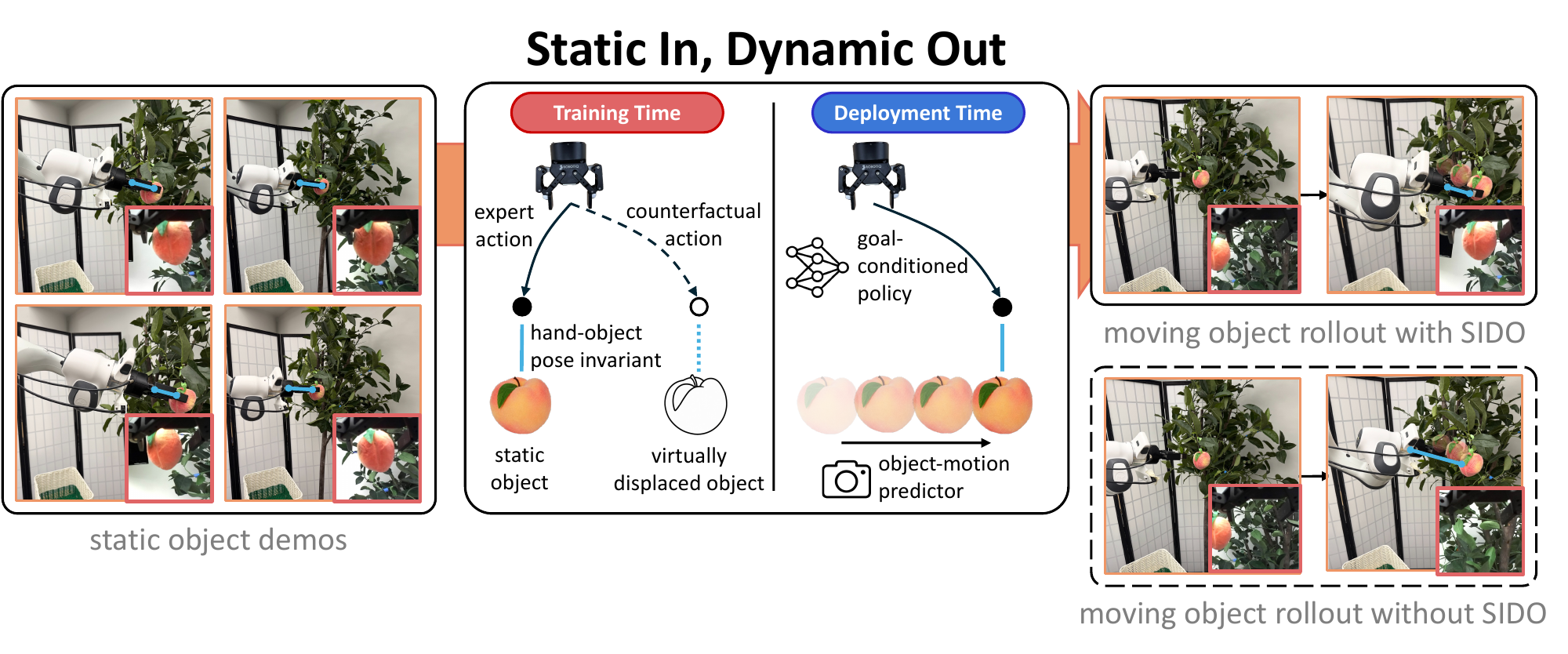}
    \vspace{-10pt}
    \caption{\textbf{Static In, Dynamic Out (SIDO)} enables a policy trained only on static object demonstrations to manipulate moving objects. At training time, SIDO creates counterfactual data pairs by virtually displacing the object and morphing the expert action to preserve the hand-object relative pose, and trains a goal-conditioned policy on them. At deployment, an object pose predictor supplies the object's future pose, which the policy reaches for. SIDO successfully grasps the moving object, while the baseline commits to the object's past pose and misses.}
    \label{fig:SIDO_teaser}
    \vspace{-10pt}
\end{figure}

Imitation learning has become a standard recipe for visuomotor manipulation, with diffusion- and transformer-based policies solving long-horizon, contact-rich tasks from demonstrations~\cite{chi2024diffusionpolicyvisuomotorpolicy, zhao2023learningfinegrainedbimanualmanipulation,reuss2023goal}. Their success rests on training data that covers the deployment distribution, and a rarely stated assumption is that the target object stays static during robot execution. Real deployments violate this constantly: parts drift on conveyors and fruits sway in the wind. Collecting new teleoperated demonstrations on moving objects is expensive, often unsafe, and limited to the motion profiles seen during collection, whereas large-scale static object demonstration datasets are already available across the community~\cite{embodimentcollaboration2025openxembodimentroboticlearning, khazatsky2025droidlargescaleinthewildrobot}. We therefore ask: \emph{can a policy trained only on static object demonstrations manipulate moving objects whose motion was unseen during training?}

Prior work touches this question without resolving it. Reactive policies~\cite{xie2026dynamicvlavisionlanguageactionmodeldynamic, black2025realtimeexecutionactionchunking, arachchige2025sailfasterthandemonstrationexecutionimitation, zhang2024catchitlearningcatch, xue2025reactivediffusionpolicyslowfast} shorten perception-to-execution latency but assume the training data already covers the motion regime of interest. Counterfactual data augmentation synthesizes samples along spatial or semantic axes~\cite{ameperosa2025rocodacounterfactualdataaugmentation, glossop2025castcounterfactuallabelsimprove}, not the forward-in-time object displacement a moving target requires. Equivariant policies~\cite{wang2024equivariantdiffusionpolicy, zhu2025se3equivariantdiffusionpolicyspherical} encode spatial symmetry, addressing static pose generalization rather than the temporal motion a moving scene exhibits. Model-based dynamic grasping pipelines~\cite{fang2023anygrasp, akinola2021dynamic, marturi2019dynamic, allen1993automated} reach moving objects, but they target the grasp itself and leave post-grasp behavior, such as placing, to separately engineered modules.

Our key observation is that manipulation on moving objects factorizes into two sub-problems: (i) where the object will be one chunk-horizon ahead, a pose-forecasting problem that needs only object trajectories with no teleoperation, and (ii) given that future object position, what action chunk achieves the correct hand-object relative pose, a goal-conditioned reaching problem the static object demonstrations already cover densely. Decoupling these into separate modules lets each be trained from data cheap to collect in isolation, without moving object teleoperation.

We introduce \textbf{\textsc{Static In, Dynamic Out (SIDO)}}, a counterfactual action augmentation that applies to any policy class with an object-pose input and pairs plug-and-play with any object pose predictor. For each static training transition, we displace the object to a counterfactual future position and morph the action chunk so its hand-object relative pose at $T_a$ and $T_p$ matches the expert demonstration, producing a goal-conditioned policy from purely static data. At deployment, a lightweight object pose predictor supplies the future object position at each chunk boundary, and can be swapped per deployment regime without retraining the policy. On three simulated tasks (Mug, Square, Stack) across five object motion patterns, SIDO achieves higher moving object success than the baselines while preserving static object performance, with consistent gains on two real-world platforms (Gantry, Peachtree). The broader takeaway: the gap between static demonstrations and dynamic deployment can be closed by recasting the dynamic problem as one of goal displacement rather than data collection.
\section{Related Work}

\paragraph{Imitation learning and covariate shift.}
Imitation learning is a common recipe for teaching robots behaviors from demonstrations~\cite{chi2024diffusionpolicyvisuomotorpolicy,reuss2023goal, zhao2023learningfinegrainedbimanualmanipulation, saxena2025matterslearninglargescaledatasets, liu2025immimiccrossdomainimitationhuman, liu2026egoengineegocentrichumanvideos, han2026video2sim2realfullstackautonomousdexterous}. Its common failure mode is covariate shift: rolled out under its own state
distribution, the policy lets small action errors compound into states absent from training. Classic remedies add corrective supervision on off-nominal states via interactive relabeling~\cite{ross2011reductionimitationlearningstructured} or noise injection~\cite{laskey2017dartnoiseinjectionrobust}. Recent methods use learned dynamics to keep execution near the expert manifold~\cite{sun2025latentpolicybarrierlearning}, or bootstrap world models in simulation for rapid real-world adaptation~\cite{levy2026simulationdistillationpretrainingworld}. A complementary line shortens the perception-to-execution loop so policies react faster to changing observations~\cite{xie2026dynamicvlavisionlanguageactionmodeldynamic, black2025realtimeexecutionactionchunking, arachchige2025sailfasterthandemonstrationexecutionimitation} or guides them with external dynamics models at inference~\cite{du2025dynaguidesteeringdiffusionpolices}. These target robot-induced error and inference speed, but not a distinct shift: the object itself may move during execution even when the robot acts correctly. A separate line confronts this by collecting large-scale dynamic demonstrations and training a dynamics-aware policy on them~\cite{fang2026generalizableroboticmanipulationdynamic}, but such data is costly to collect and limited to the motion profiles seen during collection. SIDO instead closes the gap at the training-data level, augmenting static demonstrations with counterfactual object positions so the future displacement is carried by the supervised trajectory itself, keeping the policy in-distribution for moving object rollouts.

\paragraph{Object pose prediction.}
Predicting where a moving object will be is well studied. Upstream, 6-DoF pose estimators such as FoundationPose~\cite{wen2024foundationposeunified6dpose} give instantaneous object pose from images and can be chained into a pose history. Downstream, that history drives either a learned forecaster, such as a flow-equivariant recurrent network~\cite{keller2025flowequivariantrecurrentneural}, or a classical filter propagating a low-dimensional motion state~\cite{kalman1960new}. Any such combination can supply the future object pose moving object manipulation needs, and SIDO is agnostic to the choice: the policy is trained once and paired at deployment with whichever predictor fits the regime, without retraining.

\paragraph{Data augmentation and structure priors.}
Synthetic data generation expands the training distribution without new demonstrations. Demonstration-synthesis methods enlarge a small expert corpus into a large training set~\cite{mandlekar2023mimicgendatagenerationscalable, xue2025demogen, garrett2024skillmimicgenautomateddemonstrationgeneration}, while counterfactual relabeling produces samples along language~\cite{glossop2025castcounterfactuallabelsimprove} or rigid-body pose transforms~\cite{ameperosa2025rocodacounterfactualdataaugmentation}. Closest to our setting, CCIL~\cite{ke2024ccilcontinuitybaseddataaugmentation} generates counterfactual actions implicitly through a learned dynamics model. But since static object demonstrations contain no object motion, the relevant dynamics cannot be learned implicitly and must be supplied as an explicit prior. Orthogonally, equivariant policies encode geometric structure into the model, improving sample efficiency and pose generalization under spatial transformations rather than temporal motion~\cite{wang2024equivariantdiffusionpolicy, tie2025etseedefficienttrajectorylevelse3, zhu2025se3equivariantdiffusionpolicyspherical, yang2024equibotsim3equivariantdiffusionpolicy}. SIDO supplies the geometric-temporal counterfactual these methods lack, using an explicit prior of its own: the hand-object relative pose is held invariant as each static action chunk is re-anchored to the displaced object.

\paragraph{Model-based dynamic grasping.}
Model-based pipelines grasp moving objects without demonstrations by pairing motion prediction with grasp and arm planning, closed-loop grasp detection, or temporal grasp tracking~\cite{fang2023anygrasp, akinola2021dynamic, marturi2019dynamic, allen1993automated, morrison2018closing}. They synthesize grasps rather than complete tasks, leaving post-grasp behavior to separate modules. SIDO instead learns grasping and post-grasp behavior in a single policy from static demonstrations.
\section{Problem Statement}
\label{sec:problem_statement}

\paragraph{Setup.}
Let $P_t \in \mathbb{R}^3$ denote the position of the task-relevant object at time $t$. The robot proprioception is $s_t = (p_t^{\text{eef}},\, \omega_t^{\text{eef}},\, g_t)$, stacking end-effector position, orientation, and gripper state. The robot observation $o_t$ assembles an agent view image $I_t^{\text{agent}}$, a wrist view image $I_t^{\text{wrist}}$, $s_t$, and the object position $P_t$ through an observation function $\phi$. A behavior-cloning policy $\pi_\theta : \mathcal{O} \to \mathcal{A}^{T_p}$ maps $o_t = \phi\bigl(I_t^{\text{agent}},\, I_t^{\text{wrist}},\, s_t,\, P_t\bigr)$ to a chunk of $T_p$ actions, $\hat{a}_{t:t+T_p-1} = \pi_\theta(o_t)$, where $T_p$ is the prediction horizon. Each action $a_t$ is an end-effector command, either an absolute pose or a relative pose increment, paired with a continuous gripper command in $[0,1]$. Executing $a_{t}$ advances the position of the end-effector to $p_{t+1}^{\text{eef}}$, so a chunk drives the end-effector along $p_t^{\text{eef}} \to p_{t+T_p}^{\text{eef}}$. The first $T_a \le T_p$ actions of each chunk are executed before re-planning, where $T_a$ is the action horizon.

\paragraph{Training and Deployment.}
We are given $N$ static object demonstrations $\mathcal{D}_{\text{train}} = \{(o_t^{(i)}, a_t^{(i)}) : i \in \{1, \dots, N\},\, t \in \{0, \dots, T^{(i)}-1\}\}$ with $P_t^{(i)} = P^{(i)}_{\text{static}}$ for all $t \le T_g^{(i)}$, where $T^{(i)}$ is the length of demonstration $i$, $T_g^{(i)}$ is its grasp index, and $P^{(i)}_{\text{static}}$ is the object's static pre-grasp position. Behavior cloning fits $\pi_\theta$ by minimizing $\mathbb{E}_{\mathcal{D}_{\text{train}}}[\|\pi_\theta(o_t) - a_{t:t+T_p-1}\|]$. At deployment the object position is no longer fixed: it follows an unknown trajectory $\xi : \mathbb{R}_{\ge 0} \to \mathbb{R}^3$ drawn from a deployment distribution $\Xi$ unseen during training, so $P_t = \xi(t)$. Both camera views now depend on $\xi(t)$, and the deployment observation $o_t^{\text{dep}} = \phi\bigl(I_t^{\text{agent,dep}}, I_t^{\text{wrist,dep}}, s_t^{\text{dep}}, \xi(t)\bigr)$ generally falls outside $\operatorname{supp}(\mathcal{D}_{\text{train}})$. 
Behavior cloning then fails for two compounding reasons. \emph{(1) Action mismatch:} a successful chunk committed at time $t$ must land the gripper according to the object's future position $\xi(t+T_p)$ at the chunk terminus, not its current position $\xi(t)$, since the object keeps moving during execution, but static demonstrations contain no such anticipatory actions. \emph{(2) Out-of-distribution observation:} a stale chunk lands the gripper where the object was, so the next observation drifts further from $\operatorname{supp}(\mathcal{D}_{\text{train}})$.

\section{Method}
\label{sec:method}

\vspace{-10pt}

\begin{figure}[ht]
    \centering
    \includegraphics[width=0.95\textwidth]{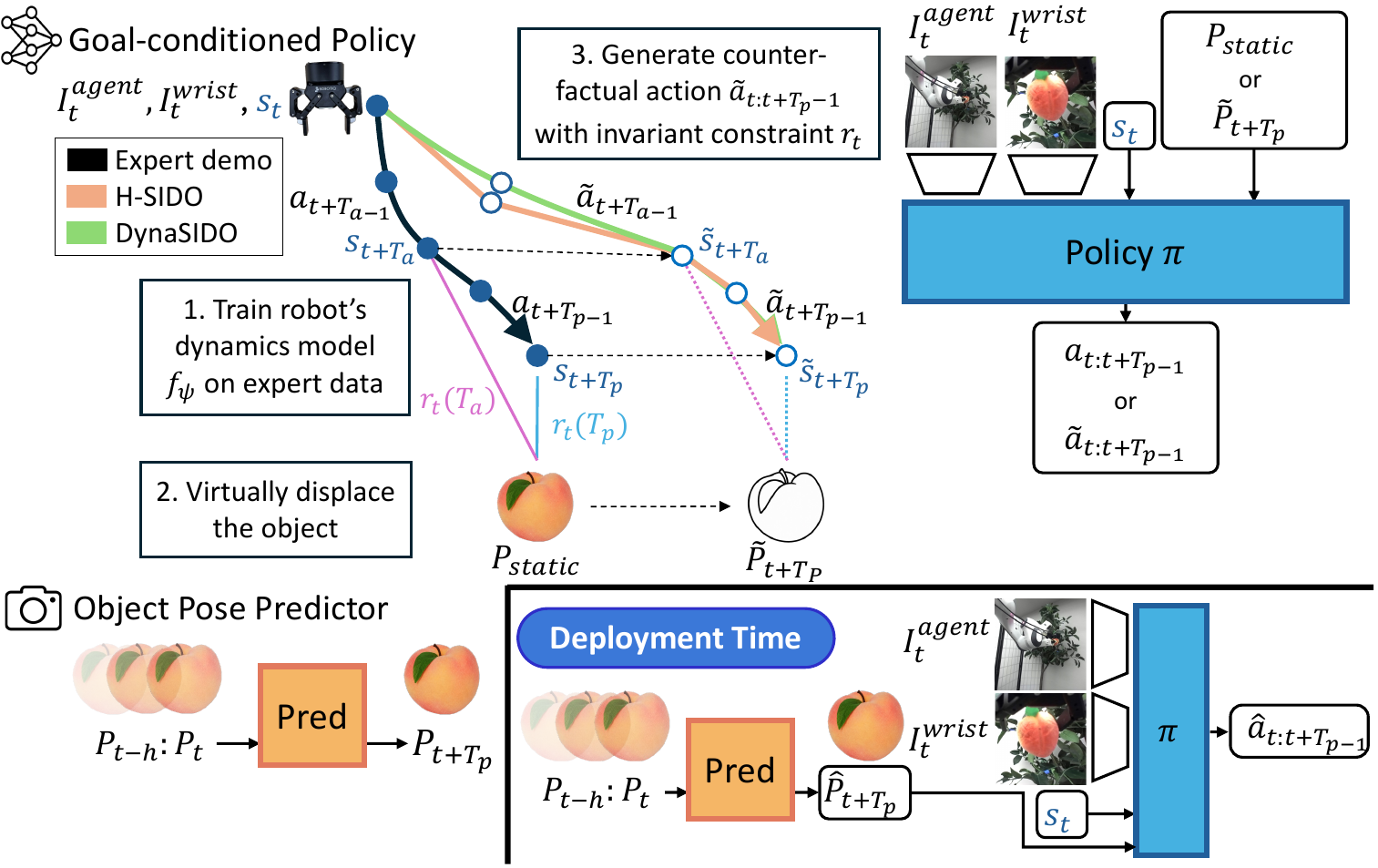}
    \caption{\textbf{SIDO overview.} \emph{Goal-conditioned policy:} from each static demonstration, SIDO (1) fits a robot dynamics model $f_\psi$ on expert data, (2)  virtually displaces the object, and (3) morphs the expert chunk into a counterfactual chunk $\tilde{a}_{t:t+T_p-1}$ that preserves the hand-object relative pose at $T_a$ and $T_p$. H-SIDO uses a heuristic ramp; DynaSIDO refines it under $f_\psi$. \emph{Object pose predictor:} SIDO pairs plug-and-play with any object pose predictor, which forecasts the object's future position $\hat{P}_{t+T_p}$ from an object pose history $P_{t-h}{:}P_t$. \emph{Deployment:} the predictor supplies the predicted future object position, which conditions the policy $\pi$ to produce the action chunk.}
    \label{fig:SIDO_method}
    \vspace{-15pt}
\end{figure}

We split moving object manipulation into two modules: a \emph{goal-conditioned policy} and an \emph{object pose predictor}. \S\ref{sec:method-invariant} defines the hand-object relative pose and the invariant constraint the policy must satisfy. \S\ref{sec:method-cf} generates counterfactual observation-action pairs from static demonstrations and trains the goal-conditioned policy on them. \S\ref{sec:method-predictor} describes the object pose predictor, which supplies the object's future position at deployment.

\subsection{Hand-Object Invariant}
\label{sec:method-invariant}

For an end-effector position $p_t^{\text{eef}} \in \mathbb{R}^3$ and object position $P_t$, define the hand-object relative pose at offset $k$ as
\begin{equation}
    r_t(k) = p_{t+k}^{\text{eef}} - P_{t+k},
    \qquad k \in \{0, \dots, T_p\}.
    \label{eq:rel_pose}
\end{equation}
In a static object demonstration, $P_{t+k} = P_{\text{static}}$ throughout the pre-grasp window $t \le T_g$, so $r_t(k)$ is the gripper's offset from the object, invariant to where the chunk lies in the world. The policy depends on this offset at two horizons: $r_t(T_a)$, the end-effector state reached when it re-plans, and $r_t(T_p)$, the state at the chunk terminus. Our augmentation preserves $r_t(T_a)$ and $r_t(T_p)$, and we call this the
\emph{invariant constraint}.

\subsection{Counterfactual Augmentation}
\label{sec:method-cf}

For each transition $(o_t,\, a_{t:t+T_p-1})$ with $t + T_p \le T_g$, we sample a displacement and produce a counterfactual pair $(\tilde{o}_t,\, \tilde{a}_{t:t+T_p-1})$ satisfying the invariant constraint.

\paragraph{Counterfactual displacement and observation.}
We draw a uniform direction $\theta \sim \mathcal{U}[0, 2\pi)$ and set $\delta = v\, T_p\, [\cos\theta,\, \sin\theta,\, 0]^\top$, where $v$ is a free parameter of the augmentation controlling the displacement magnitude, so $v\,T_p$ is the displacement accumulated over the prediction horizon. With probability $\alpha$ the sample is returned unchanged to preserve static object performance. Otherwise we displace the object by $\delta$ to the counterfactual future position $\tilde{P}_{t+T_p} \triangleq P_{\text{static}} + \delta$. By construction the policy reads the object's position from the dedicated observation channel, not from the pixels, so a valid counterfactual need only edit that channel: we keep the images $I_t^{\text{agent}}$ and $I_t^{\text{wrist}}$ as-is and write the displaced position into the pose channel, giving $\tilde{o}_t = \phi(I_t^{\text{agent}}, I_t^{\text{wrist}}, s_t, \tilde{P}_{t+T_p})$. This mismatch between the unchanged images and the displaced pose channel mirrors deployment, where the predictor writes the object's future position into that same channel.

\paragraph{Heuristic counterfactual action generation (H-SIDO).}
H-SIDO morphs the demonstrated chunk to satisfy the invariant constraint against the displaced object at $\tilde{P}_{t+T_p}$, spreading $\delta$ across the chunk with a ramp. For absolute actions, each $a_{t+k}$ is a world-frame waypoint, so we shift each waypoint directly,
\begin{equation}
    \tilde{a}_{t+k} = a_{t+k} + \rho_k\, \delta,
    \qquad
    \rho_k = \frac{\min(k,\, T_a - 1)}{T_a - 1},
    \qquad k \in \{0, \dots, T_p - 1\},
    \label{eq:heuristic_cf}
\end{equation}
where $\rho_k \in [0,1]$ rises linearly from $\rho_0 = 0$ and holds at $1$ for $k \ge T_a - 1$. For relative actions, each $a_{t+k}$ is a step increment, so we apply the same ramp to the cumulative path, indexed by arc length to preserve the trajectory shape. The closed form is in Appendix~\ref{app:heuristic_cf_rel}.

\paragraph{Dynamics-aware counterfactual action generation (DynaSIDO).}
Equation~\eqref{eq:heuristic_cf} preserves the invariant constraint on the commanded waypoints, but the closed-loop controller may track a different end-effector trajectory once the chunk is executed. DynaSIDO refines the chunk so the executed trajectory hits the displaced targets.

\emph{Dynamics model.} We fit a residual MLP $f_\psi$ on the expert transitions in $\mathcal{D}_{\text{train}}$. From the proprioception
$s_t = [\,p_t^{\text{eef}}, \omega_t^{\text{eef}}, g_t\,]$ augmented with finite-difference velocity and acceleration, $\bar{s}_t = [\,s_t,\, \dot{p}_t^{\text{eef}},\, \ddot{p}_t^{\text{eef}}\,]$, the model predicts the next state $\hat{s}_{t+1} = s_t + f_\psi(\bar{s}_t,\, a_t)$. Since the object never moves during pre-grasp, $f_\psi$ captures only the robot's controller dynamics. Rolling $f_\psi$ forward from $s_t$ under a candidate chunk $a_{t:t+T_p-1}$ yields a predicted end-effector trajectory $\hat{p}^{\text{eef}}_{t:t+T_p}$.

\emph{Targets and MPPI refinement.} The invariant constraint fixes the gripper's required positions at the two horizons. Since $r_t(k) = p^{\text{eef}}_{t+k} - P_{\text{static}}$ for a static demonstration, re-anchoring to $\tilde{P}_{t+T_p}$ shifts each target by $\delta$, so $p^{\text{tgt}}_{t+T_a} = p^{\text{eef}}_{t+T_a} + \delta$, and $p^{\text{tgt}}_{t+T_p} = p^{\text{eef}}_{t+T_p} + \delta$. We use MPPI~\citep{williams2015modelpredictivepathintegral} to
search for a chunk whose $f_\psi$-rollout reaches these targets. For a candidate chunk $a$ with the predicted end-effector position $\hat{p}^{\text{eef}}_{t+k}(a)$ at offset $k$, the cost is
\begin{equation}
    J(a) =
    \bigl\| \hat{p}_{t+T_a}^{\text{eef}}(a) - p^{\text{tgt}}_{t+T_a} \bigr\|^2
    + \bigl\| \hat{p}_{t+T_p}^{\text{eef}}(a) - p^{\text{tgt}}_{t+T_p} \bigr\|^2.
    \label{eq:mppi_cost}
\end{equation}
The heuristic chunk from Eq.~\eqref{eq:heuristic_cf} seeds the search, and the lowest-cost sample is returned as $\tilde{a}_{t:t+T_p-1}$.

\subsection{Object Pose Predictor}
\label{sec:method-predictor}
At deployment the policy conditions each chunk on the object's predicted future position $\hat{P}_{t+T_p}$, the same channel the displacement $\delta$ occupied at training. Any module that forecasts this future position can supply it, so SIDO is decoupled from the predictor and pairs with it plug-and-play: the policy is trained once and the predictor swapped per deployment regime with no retraining.
\section{Experiments}
\label{sec:experiment}

\begin{figure*}
    \centering
    \includegraphics[width=0.95\textwidth]{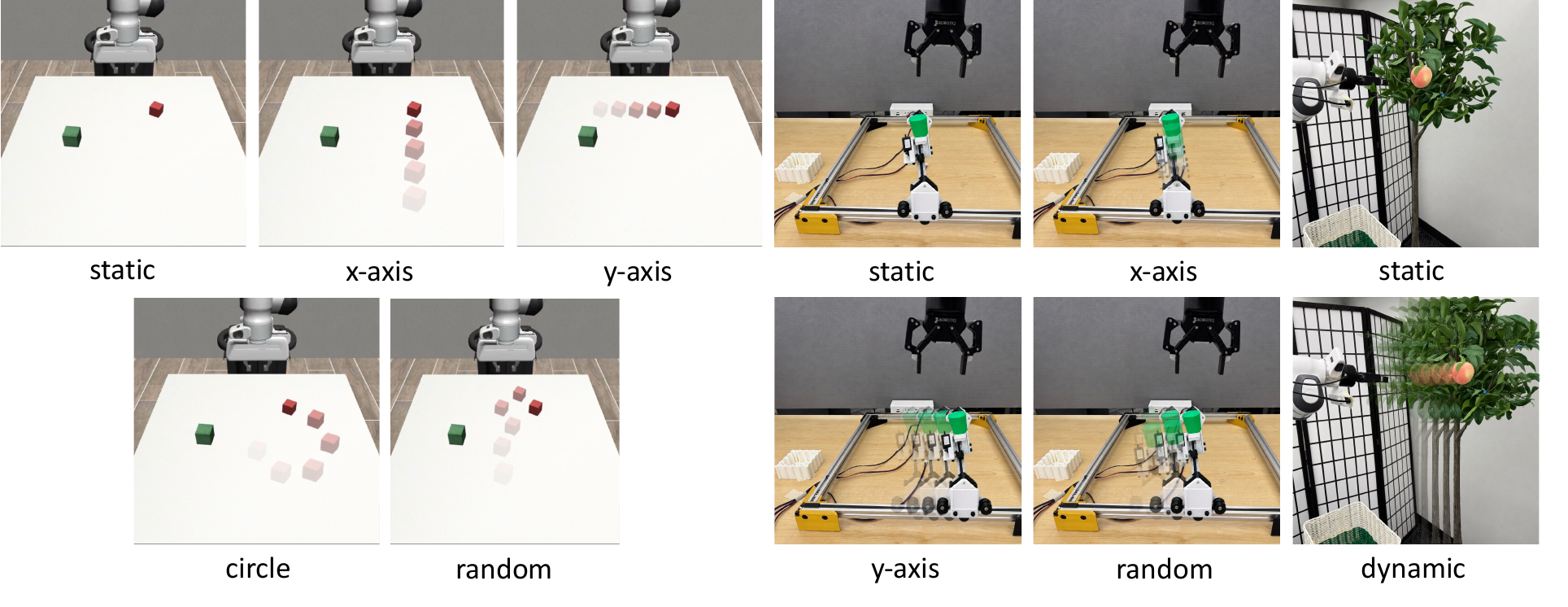} 
    \vspace{-10pt}
    \caption{\textbf{Object motion patterns across simulation and real-world tasks.}
    \emph{Left:} simulation tasks under five object motion patterns (static, x-axis, y-axis, circle, random).
    \emph{Middle:} real-world Gantry under four object motion patterns (static, x-axis, y-axis, random).
    \emph{Right:} real-world Peachtree under two object motion patterns (static, dynamic).}
    \label{fig:motion_profile}
    \vspace{-20pt}
\end{figure*}

We evaluate SIDO in simulation and the real-world, and on two policy classes to test transfer across policy classes: Diffusion Policy (DP)~\cite{chi2024diffusionpolicyvisuomotorpolicy} with a DINOv3~\cite{siméoni2025dinov3} encoder, and Equivariant Diffusion Policy (EquiDP)~\cite{wang2024equivariantdiffusionpolicy} with an equivariant ResNet18~\cite{he2015deepresiduallearningimage, e2cnn, cesa2022a}. Our experiments test three hypotheses (H1--H3), each stated and answered below. We use DynaSIDO in simulation to evaluate the dynamics-aware variant and H-SIDO in our current hardware evaluation, where it empirically approximates DynaSIDO in the low-speed regime. Appendix~\ref{app:hsido_dynasido} further characterizes the benefit of DynaSIDO at higher object speeds. Hardware and motion-pattern details are in Appendices~\ref{app:hardware} and~\ref{app:motion}.

For the object pose predictor we use a finite-difference predictor by default, which extrapolates the object's position one chunk-horizon ($T_p$) ahead from the last three object pose observations. To test plug-and-play (H3), we also swap in two learned predictors on the fixed policy without retraining: an MLP that regresses the future position from the last 8 object pose observations, and a flow-equivariant recurrent network (FERNN) that builds in equivariance to object motion over the same 8-step history. Predictor details are in Appendix~\ref{app:predictors}.

\paragraph{Baselines.}
Each baseline adds one way of handling object motion, from none to test-time correction. \textbf{DP} receives image and proprioception with no object pose channel. \textbf{DP (cur. pose)} adds the object's current position as input but trains only on static data, testing whether observing the object suffices. \textbf{DP+AC} (action compensation) adds a test-time correction that shifts each commanded action by the object's current displacement, testing whether motion handling can be bolted on after training without retraining. \textbf{DP+SIDO} (ours) instead bakes the displacement into the supervised chunk at training and conditions on the predicted future pose, and pairs with a swappable predictor to test plug-and-play. The EquiDP variants follow the same pattern.

\paragraph{Tasks.}
We test the object motion patterns of Fig.~\ref{fig:motion_profile} across both simulation and real-world tasks. In simulation we build on robosuite~\cite{DBLP:journals/corr/abs-2009-12293} and robomimic~\cite{DBLP:journals/corr/abs-2108-03298} with three MimicGen~\cite{mandlekar2023mimicgendatagenerationscalable} tasks, Stack (200 demonstrations), Mug, and Square (1000 each), the object moving at $2$\,cm/s at test time. In real-world tasks we use the DROID~\cite{khazatsky2025droidlargescaleinthewildrobot} setup with DP as the policy class, the object translating at $1.5$\,cm/s at test time. \emph{Gantry:} pick a cylindrical capsule and place it in a box. We collect 30, 22, and 64 static demonstrations whose placements cover the $23$\,cm x-axis, $43$\,cm y-axis, and random workspace ranges, and evaluate each under both static and dynamic conditions. \emph{Peachtree:} pick a peach from a tree swayed by a mobile base and drop it into a basket below, over a $30$\,cm path. We collect 25 static demonstrations and evaluate under static and dynamic conditions.

\begin{table*}[ht]
    \centering
    \label{tab:sim-eval}
    \scriptsize
    \begin{tblr}{
      width   = \linewidth,
      colsep  = 0.5pt,
      rowsep  = 1pt,
      colspec = {Q[l] *{3}{X[c]X[c]X[c]X[c]X[c]}},
      row{1}  = {font=\bfseries},
      hline{1,9} = {-}{wd=1pt},
      hline{3,6} = {-}{},
      cell{1}{2}  = {c=5}{c},
      cell{1}{7}  = {c=5}{c},
      cell{1}{12} = {c=5}{c},
    }
    & Mug & & & & &
      Square & & & & &
      Stack \\

    Method
    & static & x-axis & y-axis & circle & random
    & static & x-axis & y-axis & circle & random
    & static & x-axis & y-axis & circle & random \\

    DP (cur. pose)
    & 0.20 & 0.10 & 0.00 & 0.15 & 0.15
    & \textbf{0.60} & 0.00 & 0.20 & 0.15 & 0.20
    & \textbf{1.00} & 0.15 & 0.40 & 0.35 & 0.65 \\

    DP + AC
    & -- & 0.00 & 0.10 & 0.10 & 0.10
    & -- & 0.30 & 0.00 & 0.35 & 0.20
    & -- & 0.15 & 0.30 & 0.40 & 0.60 \\

    \textbf{DP + SIDO}
    & \textbf{0.25} & \textbf{0.35} & \textbf{0.20} & \textbf{0.20} & \textbf{0.30}
    & \textbf{0.60} & \textbf{0.50} & \textbf{0.40} & \textbf{0.40} & \textbf{0.55}
    & 0.95 & \textbf{0.90} & \textbf{0.65} & \textbf{0.65} & \textbf{0.95} \\

    EquiDP (cur. pose)
    & 0.35 & \textbf{0.35} & \textbf{0.05} & \textbf{0.05} & \textbf{0.15}
    & \textbf{0.60} & 0.10 & 0.05 & 0.05 & 0.10
    & \textbf{0.95} & 0.35 & 0.55 & 0.25 & 0.65 \\

    EquiDP + AC
    & -- & \textbf{0.35} & \textbf{0.05} & \textbf{0.05} & \textbf{0.15}
    & -- & 0.05 & 0.00 & 0.05 & \textbf{0.20}
    & -- & 0.35 & 0.55 & 0.65 & 0.90 \\

    \textbf{EquiDP + SIDO}
    & \textbf{0.50} & 0.10 & \textbf{0.05} & 0.00 & 0.00
    & \textbf{0.60} & \textbf{0.20} & \textbf{0.10} & \textbf{0.20} & \textbf{0.20}
    & \textbf{0.95} & \textbf{0.70} & \textbf{0.70} & \textbf{0.75} & \textbf{0.95} \\

    \end{tblr}
    \caption{\textbf{Simulation success rate ($\uparrow$)} on three MimicGen tasks under five object motion patterns, 20 rollouts per cell.}
    \label{tab:sim-eval}
    \vspace{-10pt}
\end{table*}

\begin{table}[ht]
\centering
\begin{minipage}[t]{0.62\linewidth}
    \centering
    \label{tab:gantry-eval}
    \scriptsize
    \begin{tblr}{
      width   = \linewidth,
      colsep  = 0.5pt,
      rowsep  = 1pt,
      colspec = {Q[l] *{3}{X[c]X[c]}},
      row{1}  = {font=\bfseries},
      hline{1,7} = {-}{wd=1pt},
      hline{2} = {2-7}{},
      hline{3} = {-}{},
      cell{1}{2} = {c=2}{c},
      cell{1}{4} = {c=2}{c},
      cell{1}{6} = {c=2}{c},
    }
    & x-axis & & y-axis & & random \\
    Method
    & static & dynamic
    & static & dynamic
    & static & dynamic \\
    
    DP
    & \textbf{0.90} & 0.10
    & 0.85 & 0.35
    & 0.80 & 0.15 \\
    
    DP (cur. pose)
    & 0.80 & 0.35
    & \textbf{0.95} & 0.50
    & 0.90 & 0.20 \\
    
    DP + AC
    & -- & 0.00
    & -- & 0.30
    & -- & 0.00 \\
    
    \textbf{DP + SIDO}
    & 0.80 & \textbf{0.55}
    & \textbf{0.95} & \textbf{0.65}
    & \textbf{1.00} & \textbf{0.25} \\
    
    \end{tblr}
    \vspace{5pt}
    \caption{\textbf{Real-world Gantry success rate ($\uparrow$)} under static and dynamic conditions for three motion patterns, 20 rollouts per cell.}
    \label{tab:gantry-eval}
\end{minipage}
\hfill
\begin{minipage}[t]{0.36\linewidth}
    \centering
    \label{tab:peachtree-eval}
    \scriptsize
    \begin{tblr}{
      width = \linewidth,
      colsep = 4pt,
      colspec = {Q[l] X[c] X[c]},
      row{1} = {font=\bfseries},
      hline{1,6} = {-}{wd=1pt},
      hline{2} = {-}{},
    }

    Method & static & dynamic \\

    DP & 0.65 & 0.30 \\
    DP (cur. pose) & \textbf{0.90} & 0.45 \\
    DP + AC & -- & 0.30 \\
    \textbf{DP + SIDO} & 0.75 & \textbf{0.60} \\

    \end{tblr}
    \vspace{5pt}
    \caption{\textbf{Real-world Peachtree success rate ($\uparrow$)} under static and dynamic conditions, 20 rollouts per cell.}
    \label{tab:peachtree-eval}
\end{minipage}
\vspace{-25pt}
\end{table}

\begin{figure*}[ht]
    \centering
    \includegraphics[width=0.95\textwidth]{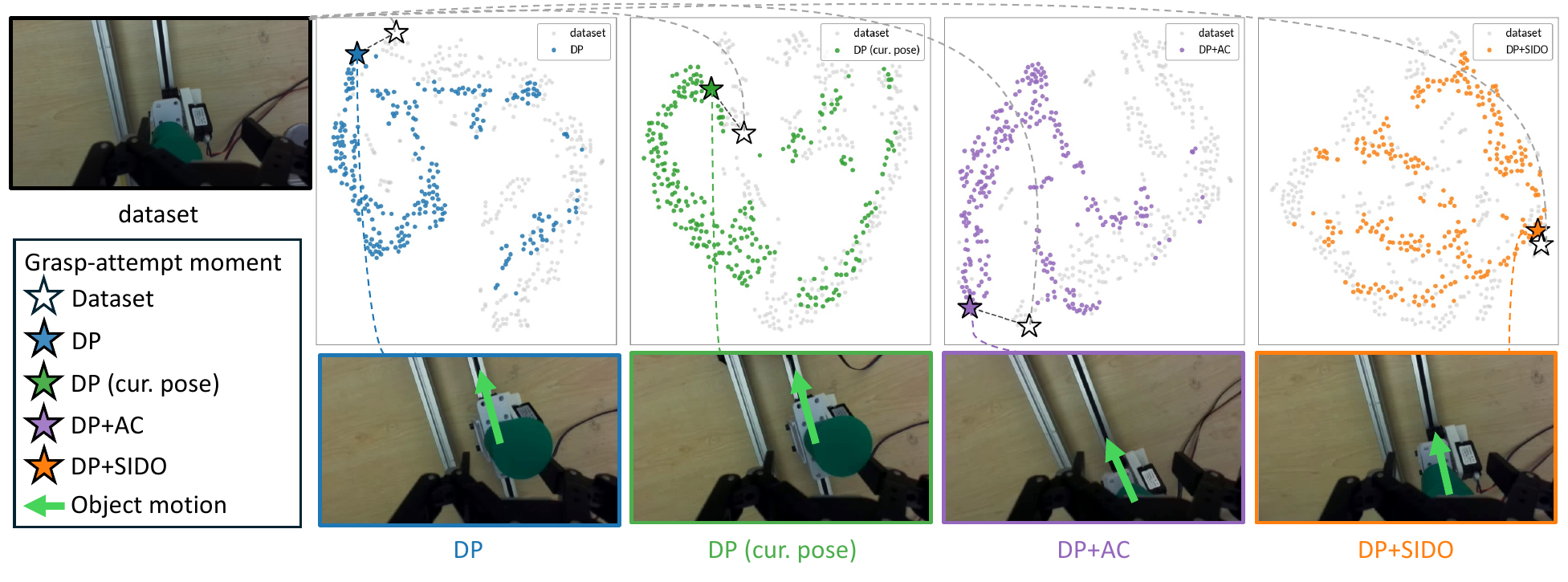}
    \caption{\textbf{SIDO's wrist view stays in-distribution with the static
    training data.} t-SNE of pre-grasp wrist frames in each method's
    fine-tuned DINOv3 space, 20 rollouts per method on Gantry x-axis.
    Bottom row: wrist view at the grasp-attempt moment. SIDO's frame matches
    the dataset reference, while baselines drift away and miss or off-center
    the object.}
    \label{fig:tsne}
    \vspace{-15pt}
\end{figure*}

\paragraph{H1: SIDO improves moving object success while preserving static object
performance.}
Across simulation (Table~\ref{tab:sim-eval}) and the real-world tasks (Tables~\ref{tab:gantry-eval},~\ref{tab:peachtree-eval}), SIDO improves over the baselines on moving objects while matching or nearly matching them on static objects, lifting Gantry x-axis dynamic to $0.55$ from the best baseline's $0.35$ and Peachtree dynamic to $0.60$ from $0.45$. The baseline ladder isolates why. With no object pose channel, DP collapses to $0.10$ on Gantry x-axis dynamic; the current pose raises this to $0.35$ but no further, since the policy sees the object yet was never trained to act on its motion. AC supplies that motion as a test-time waypoint shift, but the correction lands after the policy commits, so the controller smooths it away rather than tracking it, falling to $0.00$, below even plain DP. SIDO's gain is largest along the wrist-camera x-axis: a stale chunk lets the object slide out of the wrist view, forcing the policy onto a frame unlike any in training, whereas SIDO reaches for the predicted position and keeps the object in view, so each grasp attempt stays in distribution (Fig.~\ref{fig:tsne}). This is sharpest on Stack x-axis, where SIDO lifts success from $0.15$ to $0.90$. The exception is EquiDP+SIDO on Mug, where the gap between the commanded and executed trajectories is enough to miss a precision-critical grasp even though SIDO commands a pose that tracks the object closely.

SIDO also preserves static object performance, matching the baselines on static objects and topping them on Gantry random at $1.00$. With probability $\alpha$ (\S\ref{sec:method-cf}) the counterfactual sample is returned unchanged, so the static distribution is never abandoned, mirroring PUMA~\cite{fang2026generalizableroboticmanipulationdynamic}'s finding: mixing static demonstrations into dynamic training helps because static data anchors foundational manipulation while the dynamic component supplies temporal variation. SIDO realizes this mixing with no dynamic demonstrations, the counterfactual augmentation supplying the temporal component.

\paragraph{H2: SIDO preserves the hand-object relative pose at the action
terminus.}
We test whether the invariant constraint transfers to inference: does the policy shift its commanded gripper with the moving object as it was trained to? If so, the action terminus holds the demonstrated relative pose and the next observation stays in distribution.

\begin{figure*}[ht]
    \vspace{-10pt}
    \centering
    \includegraphics[width=\textwidth]{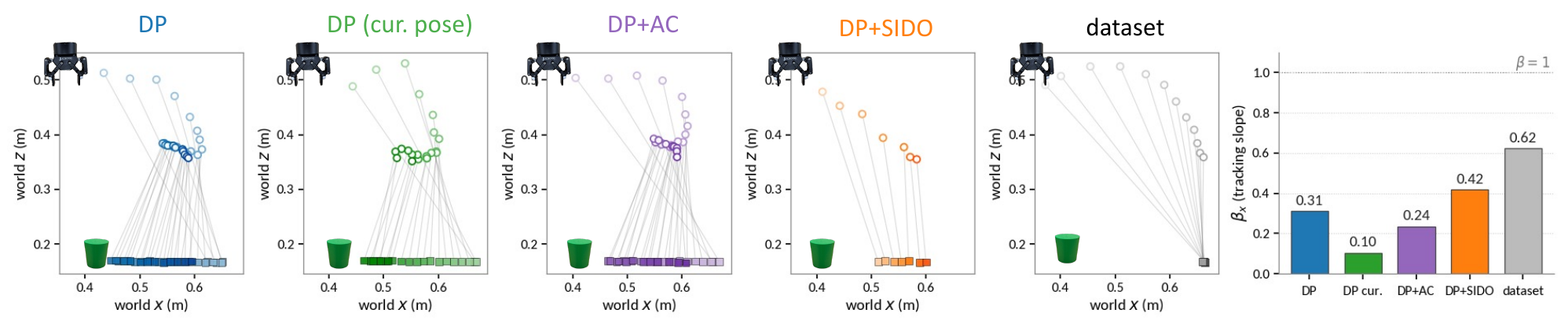}
    \caption{\textbf{Gripper tracks the object under SIDO, baselines lag behind.} World-frame $x$-$z$ scatter of commanded EEF at action terminus $T_a$ (circles) and object position (squares) on Gantry x-axis. SIDO's commands sweep with the object, preserving the hand-object relative pose, baselines do not. \emph{Right:} tracking slope $\beta_x$ of commanded EEF on observed object $x$, 20 rollouts.}
    \label{fig:tracking}
    \vspace{-10pt}
\end{figure*}

SIDO's commanded gripper tracks the moving object far closer than any baseline.  At each terminus $T_a$ we take the commanded end-effector and observed object positions along the wrist-camera $x$-axis, $p^{\text{eef}}_{\text{cmd},x}$ and $P_x$, pooled over all 20 rollouts on Gantry x-axis (successes and failures), and fit the least-squares slope $\beta_x = \mathrm{Cov}(p^{\text{eef}}_{\text{cmd},x},\, P_x) / \mathrm{Var}(P_x)$, where $\beta_x = 1$ tracks the object one-for-one and $\beta_x = 0$ ignores its motion (Fig.~\ref{fig:tracking}). The static demonstrations, differing only in object placement, set an expert-demonstration reference of $\beta_x = 0.62$. DP+SIDO comes closest at $0.42$. The baselines trail at DP $0.31$, DP+AC $0.24$, and DP (cur. pose) $0.10$. DP (cur. pose) sitting lowest confirms the current pose alone is not enough: with no motion signal the gripper lags, whereas SIDO's augmentation writes that motion into the supervision and the gripper tracks the object at rollout.

This tracking breaks the compounding loop in which an action mismatch drives the observation out of distribution. Fig.~\ref{fig:tsne} shows t-SNE embeddings of pre-grasp wrist frames in each method's fine-tuned DINOv3 encoder over the same rollouts: DP+SIDO's frames overlap the static training distribution and its grasp-attempt view matches the training reference, while the baselines lag the object, drift to a separate region of the encoder space, and miss.

\paragraph{H3: SIDO transfers across policy classes and predictor choices.}
SIDO applies across policy classes. 
Table~\ref{tab:sim-eval} shows that SIDO is the strongest DP variant across all four moving-object patterns on all three tasks, and the strongest EquiDP variant on Square and Stack. SIDO can also pair with a plug-and-play predictor without retraining the policy. 
Table~\ref{tab:predictors_main} reports DP+SIDO on Gantry random with three predictors (finite difference, MLP, and FERNN). We collect five minutes of moving object trajectories and train each predictor on them. Among the three, FERNN attains the lowest forecast error and the highest success of $0.40$, the gain coming from a velocity estimate that re-acquires quickly after each direction change in the piecewise-constant random motion. Beyond task success, we report each predictor's forecast accuracy as the Final Displacement Error (FDE): the Euclidean distance between the predicted object position one chunk-horizon ($T_p$) ahead and the object's observed position at that step, averaged over the pre-grasp window. From this we find that the predictor with the lowest forecast error also achieves the highest success rate.

\begin{wraptable}{r}{0.45\textwidth}
    \centering
    \vspace{-12pt}
    \begin{tabular}{l c c}
    \toprule
    Predictor & SR$\uparrow$ & FDE ($\downarrow$, cm) \\
    \midrule
    Finite difference & 0.25 & 2.97 \\
    MLP               & 0.25 & 2.21 \\
    FERNN             & 0.40 & 1.77 \\
    \bottomrule
    \end{tabular}
    \caption{\textbf{Predictor swap on real-world Gantry (random).} Success rate (SR) and forecast accuracy (FDE) for DP+SIDO with three object pose predictors, policy held fixed.}
    \vspace{-20pt}
    \label{tab:predictors_main}
\end{wraptable}
\section{Conclusion}

\label{sec:conclusion}

We introduce \textsc{Static In, Dynamic Out} (SIDO), a counterfactual action augmentation that trains on static object demonstrations and rolls out on moving objects. By morphing each expert chunk to hold the hand-object relative pose fixed at both the replanning point and the chunk terminus, SIDO converts static demonstrations into supervision that anticipates the object's future pose, with H-SIDO supplying a heuristic ramp and DynaSIDO refining it under a learned dynamics model. Across three simulated and two real-world tasks, SIDO improves moving object success while preserving static object performance. The preserved invariant transfers to rollout, where the commanded gripper tracks the moving object rather than lagging behind it, and the gains hold across policy classes and object pose predictors swapped at deployment with no policy retraining. Together these show that the temporal variation a moving target demands can be supplied by an explicit geometric prior at training time, leaving static demonstrations as a practical foundation for moving object manipulation without moving object teleoperation.

\section{Limitation}
\label{sec:limitation}

Our real-world pipeline relies on an off-the-shelf 6-DoF estimator, which needs a per-object reference, a CAD model or a short scan, before it can track a new object. This setup cost and any estimator error lie outside SIDO, which is agnostic to the estimator, so a more capable one improves the system without changing the method. Because SIDO edits an object-pose input, it applies to policies that expose such a channel, while pixel-only policies would require extending the counterfactual edit to the images. SIDO also translates each demonstrated reach toward the displaced target without explicitly accounting for obstacles along the shifted path. This is fine in the open tabletop scenes we test, but in clutter the morphed reach could pass close to another object, where a collision-aware morph or a short planner would extend SIDO.


\clearpage



\bibliography{references}  
\clearpage
\section{Appendix}
\label{sec:appendix}

\appendix

\section{H-SIDO Approximates DynaSIDO in the Low Speed Regime, DynaSIDO Scales Beyond It}
\label{app:hsido_dynasido}

We study how the two counterfactual action generators compare as object speed grows. We run two tests on the Stack x-axis task. The first is a chunk level comparison at the low deployment speed, measuring whether each generator lands the gripper at the invariant constraint. The second is a sweep across object speeds, measuring end-to-end task success. The first shows the two are approximately the same in the low speed regime, and the second shows DynaSIDO pulling ahead as the object moves faster.

\paragraph{H-SIDO approximates DynaSIDO at low speed.}
We compare the two generators, H-SIDO and DynaSIDO, at low speed in two ways, before training and after.

\emph{Generator level.} We sample 200 eligible pre-grasp chunks from the Stack task demonstrations and run each through both generators in closed-loop MuJoCo\cite{6386109} from a shared pre-grasp snapshot. For each chunk the object is first teleported by the counterfactual displacement $\delta$ corresponding to the $2$\,cm/s deployment speed. H-SIDO and DynaSIDO then morph the demonstrated action against the displaced object, and we execute the morphed chunk on the gripper to see whether it can track the commanded action and hold the hand-object invariant constraint at the displaced target. Table~\ref{tab:hsido_dynasido} reports two quantities. The realized mean hand-object distance $\|r_t(k)\|$ at $T_a$ and $T_p$ measures whether the executed gripper lands at the invariant constraint, and SPARC smoothness (Appendix~\ref{app:sparc}) measures execution quality. H-SIDO and DynaSIDO both stay within 2 mm of the static demonstration at both horizons, and their SPARC smoothness is comparable, with mean differences well below the per-chunk standard deviation.

\emph{Policy level.} We train two policies, one with each generator, and roll each out for 20 rollouts at the $2$\,cm/s deployment speed on Stack x-axis. Both reach the same end-to-end task success of $0.90$. Taken together, the two tests show that in the smooth, low speed regime the heuristic ramp of H-SIDO already attains what the correction of DynaSIDO provides, both in the generated chunk and in the trained policy.

\begin{table}[ht]
    \centering
    \begin{tabular}{l c c c c}
    \toprule
     & $\|r_t(T_a)\|$ (cm) & $\|r_t(T_p)\|$ (cm) & SPARC $\uparrow$ & SR $\uparrow$ \\
    \midrule
    Static demo & 12.26 & 8.10 & $-2.172 \pm 0.300$ & -- \\
    H-SIDO      & 12.38 & 8.15 & $-2.139 \pm 0.269$ & 0.90 \\
    DynaSIDO    & 12.38 & 8.16 & $-2.140 \pm 0.271$ & 0.90 \\
    \bottomrule
    \end{tabular}
    \vspace{10pt}
    \caption{\textbf{H-SIDO vs. DynaSIDO at low speed regime ($2$\,cm/s).} The first three columns report generator-level metrics over 200 Stack pre-grasp chunks, the realized hand-object distance $\|r_t(k)\|$ at the two horizons and SPARC smoothness, both as means. SR is the trained-policy success rate over 20 rollouts. H-SIDO attains approximately the same hand-object invariant $\|r_t(k)\|$, SPARC smoothness, and success rate as DynaSIDO.}
    \label{tab:hsido_dynasido}
\end{table}

\paragraph{DynaSIDO scales beyond H-SIDO as object speed rises.}
We now vary object speed and measure end-to-end task success rate on Stack x-axis. We train two policies, one with each generator, then evaluate each as the object speed rises from $0$ to $10$\,cm/s, $20$ rollouts per cell. Faster objects translate to larger per-chunk displacement $\delta$.

\begin{figure}[ht]
\centering
\begin{minipage}[c]{0.42\linewidth}
    \centering
    \small
    \setlength{\tabcolsep}{4pt}
    \begin{tabular}{r c c}
    \toprule
    Object speed & H-SIDO SR$\uparrow$ & DynaSIDO SR$\uparrow$ \\
    \midrule
    0 & \textbf{0.95} & \textbf{0.95} \\
    2 & \textbf{0.90} & \textbf{0.90} \\
    4 & 0.20 & \textbf{0.35} \\
    6 & 0.05 & \textbf{0.25} \\
    8 & 0.05 & \textbf{0.15} \\
    10 & 0.00 & \textbf{0.05} \\
    \bottomrule
    \end{tabular}
\end{minipage}
\hfill
\begin{minipage}[c]{0.54\linewidth}
    \centering
    \includegraphics[width=\linewidth]{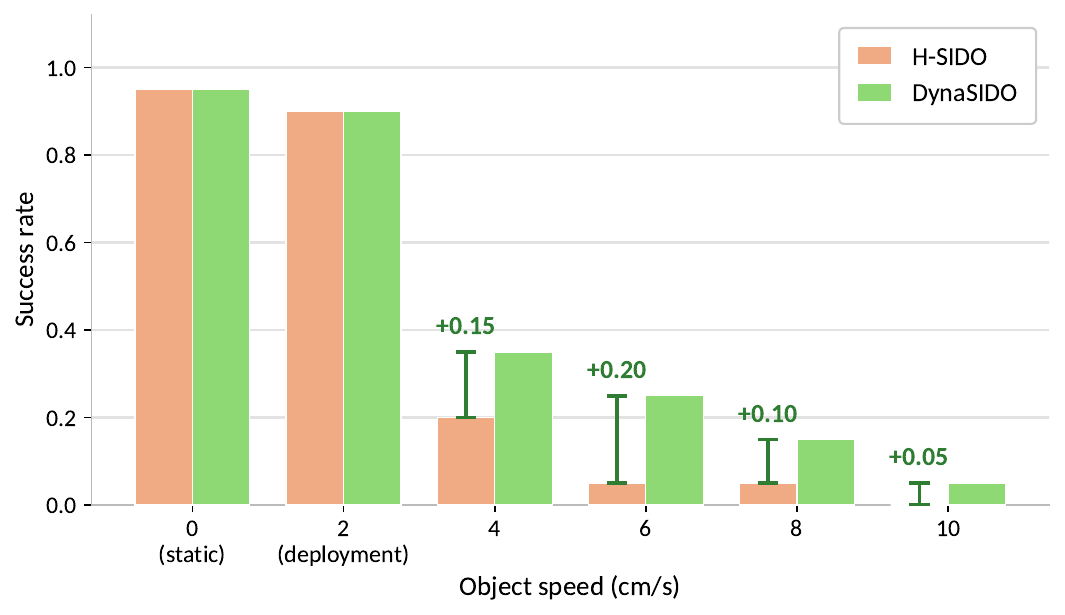}
\end{minipage}
\caption{\textbf{H-SIDO vs. DynaSIDO success rate ($\uparrow$) at higher object
speeds on Stack x-axis}, $20$ rollouts per cell. The two match at and below the
$2$\,cm/s deployment speed, and DynaSIDO pulls ahead as the object moves faster.}
\label{tab:hsido_dynasido_sweep}
\end{figure}

Table~\ref{tab:hsido_dynasido_sweep} reports the result. At and below the $2$\,cm/s the two are approximately identical, $0.95$ static and $0.90$ at $2$\,cm/s, confirming the close match of Table~\ref{tab:hsido_dynasido}. However, above it they diverge. H-SIDO's heuristic ramp spreads $\delta$ across the chunk without regard for the controller's per-step authority, so its success collapses to $0.05$ by $6$\,cm/s and to $0.00$ by $10$\,cm/s. DynaSIDO's MPPI refinement queries the dynamics model $f_\psi$ and redistributes the displacement across steps to keep the commanded chunk trackable, retaining $0.25$ at $6$\,cm/s where H-SIDO has failed. The gap peaks at $+0.20$ around $6$\,cm/s, then narrows as both approach the floor at the highest speeds.

\section{SIDO vs Training on Moving Object Demonstrations}
\label{app:dynamic_baseline}

This section compares SIDO against the coupled alternative, a single policy trained directly on moving object demonstrations.

\paragraph{Setup.} The moving object baseline conditions only on the object's current pose and does not model motion anticipation explicitly. The anticipation instead dissolves into the demonstrated actions, because teleoperating a moving object forces the operator to lead it, placing the commanded gripper where the object will be at contact rather than where it is at command time. We collect $64$ moving object demonstrations on the Gantry random task and train a DP baseline on them. SIDO trains $64$ static demonstrations of the same task and pairs with the finite difference predictor. Both are evaluated under static and dynamic conditions, $20$ rollouts per cell.

\begin{figure}[!ht]
  \centering
  \includegraphics[width=0.55\textwidth]{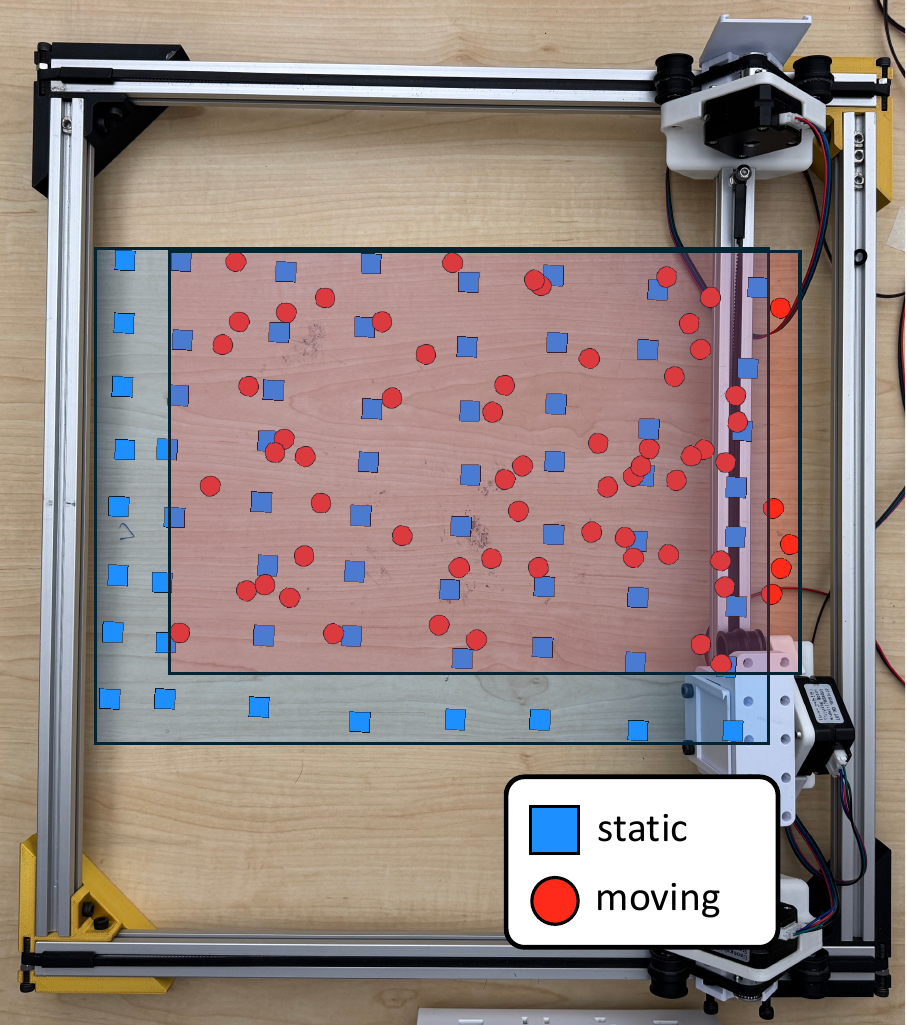}
  \caption{\textbf{Static vs moving object demonstration coverage.} Moving object reaches only $80.02\%$ of the workspace extent the static $8\times8$ grid covers.}
  \label{fig:static_vs_dynamic_coverage}
\end{figure}

\paragraph{Collection cost.} Collecting the moving object demonstrations is more expensive than the static demonstrations. Table~\ref{tab:demo_collection_cost} reports the teleoperation success rate and wall clock time to gather the same $64$ usable demonstrations on a static versus a moving object. A moving object lowers the per attempt success rate from $98.5\%$ to $59.8\%$ and lengthens collection from $1$\,h\,$49$\,m to $2$\,h\,$33$\,m, so the moving object baseline pays a markedly higher data collection cost for the same number of demonstrations. The moving object demonstrations also cover less of the workspace, reaching $80.02\%$ of the extent the static $8\times8$ grid covers (Fig.~\ref{fig:static_vs_dynamic_coverage}).

\begin{table}[ht]
    \centering
    \begin{tabular}{l c c}
    \toprule
    Demonstration dataset & Success rate $\uparrow$ & Collection time $\downarrow$ \\
    \midrule
    Static object  & $98.5\%$ & 1\,h\,49\,m \\
    Moving object  & $59.8\%$ & 2\,h\,33\,m \\
    \bottomrule
    \end{tabular}
    \vspace{6pt}
    \caption{\textbf{Cost of demonstration collection on static vs moving object.} Both datasets contain $64$ usable demonstrations. Success
    rate is the fraction of teleoperation attempts the operator marked successful.
    Collection time is wall clock from the first attempt to the last saved
    trajectory, including discarded attempts.}
    \label{tab:demo_collection_cost}
\end{table}

\paragraph{Results.} Table~\ref{tab:dynamic_baseline_results} reports the baseline trained on moving object demonstrations against DP+SIDO. The moving object baseline reaches $0.05$ on moving objects and $0.80$ on static objects. DP+SIDO, trained on the cheaper static demonstrations, reaches $0.25$ and $1.00$. SIDO uses no moving object demonstration yet surpasses the moving object baseline under both conditions, even though the moving object baseline pays the higher collection cost of Table~\ref{tab:demo_collection_cost}.

The lower static performance follows from the anticipatory offset present in moving object chunks. The policy learns to reach ahead of the object's current position by roughly its displacement over the chunk horizon. On a static object, this offset has no motion to track, so the terminus tends to land ahead of the target and the grasp can miss. SIDO reduces this effect because the anticipatory offset enters only through the counterfactual displacement, which is zeroed with probability $\alpha$ to preserve the static distribution.

Two coverage gaps explain why the moving object baseline also trails on moving objects. The first is direction. Each moving object demonstration follows a single operator trajectory, so the demonstrations span only the few motion directions collected, whereas SIDO samples the displacement direction uniformly and covers the set the test motion can take. The second is space. The moving object grasps reach only $80.02\%$ of the workspace extent the static $8\times8$ grid covers, so more test states fall outside the training support. Together these leave the moving object baseline ready for fewer motions and positions than SIDO, despite its higher collection cost. Closing both gaps would require collecting many more moving object demonstrations to span the directions and position, compounding the higher per demonstration cost.

\begin{table}[ht]
    \centering
    \begin{tabular}{l c c}
    \toprule
    Training dataset & static & dynamic \\
    \midrule
    Moving ($64$ moving object demos) & $0.80$ & $0.05$ \\
    Static + SIDO ($64$ static demos) & \textbf{1.00} & \textbf{0.25} \\
    \bottomrule
    \end{tabular}
    \vspace{10pt}
    \caption{\textbf{Static demonstrations with SIDO vs.\ moving object demonstrations success rate on Gantry random ($\uparrow$)} under static and dynamic conditions, 20 rollouts per cell.}
    \label{tab:dynamic_baseline_results}
\end{table}

\section{SIDO under Varying Object Velocity}
\label{app:varying_velocity}
This section asks whether SIDO can extend to a velocity that changes during a rollout, not just a varying direction.
\paragraph{Setup.} Two changes adapt SIDO to a range of velocities, one on each module. On the policy side, instead of fixing the displacement magnitude, we draw it uniformly from $[0, \delta]$ each time a counterfactual is sampled, so the goal-conditioned policy covers the full range of per-chunk displacements the test velocities induce rather than a single one. On the predictor side, we use FERNN, which builds in equivariance to object motion and is known to generalize across velocities, so a single predictor forecasts the future pose without a separate model per speed. We train FERNN on five minutes of Gantry random trajectories split evenly across $1.5$, $3.0$, and $4.5$\,cm/s. We then roll out DP+SIDO on Gantry random at each of the three speeds, and under a varying velocity condition that resamples the speed among the three each time a new random goal is drawn, every $31$\,cm of travel, about $20.7$\,s, $10.3$\,s, and $6.9$\,s at $1.5$, $3.0$, and $4.5$\,cm/s respectively. Each cell is $20$ rollouts.
\paragraph{Results.} Table~\ref{tab:varying_velocity} reports the four conditions. The success rate of DP+SIDO reaches $0.25$ at $1.5$\,cm/s, $0.25$ at $3.0$\,cm/s, and $0.10$ at $4.5$\,cm/s, the highest speed being the hardest. Under the varying velocity condition the policy reaches $0.40$, on par with or above the single speed conditions, so changing the object's speed mid rollout does not harm the policy. The single policy handles all four conditions without retraining, the range of training displacements covering the velocities and the FERNN forecast supplying the future pose each speed demands.

This indicates SIDO extends to varying velocity when two requirements are met: the policy is trained over a range of displacement magnitudes, and the predictor forecasts the future pose accurately as the velocity changes. Here we give broader insight: a change in object velocity is a shift of the inertial reference frame the object moves in, and SIDO absorbs it through the predicted goal and the morphed chunk rather than through new demonstrations, giving an approximate Galilean equivariance effect between the observation and the commanded action.

\begin{table}[ht]
    \centering
    \begin{tabular}{l c}
    \toprule
    Test velocity & success rate $\uparrow$ \\
    \midrule
    $1.5$\,cm/s & $0.25$ \\
    $3.0$\,cm/s & $0.25$ \\
    $4.5$\,cm/s & $0.10$ \\
    Varying ($1.5$, $3.0$, $4.5$\,cm/s) & 0.40 \\
    \bottomrule
    \end{tabular}
    \vspace{10pt}
    \caption{\textbf{DP+SIDO success rate under varying object velocity on Gantry random ($\uparrow$)}, 20 rollouts per cell.}
    \label{tab:varying_velocity}
\end{table}

\section{Hardware Setup}
\label{app:hardware}

\begin{figure}[!ht]
    \centering
    \includegraphics[width=\textwidth]{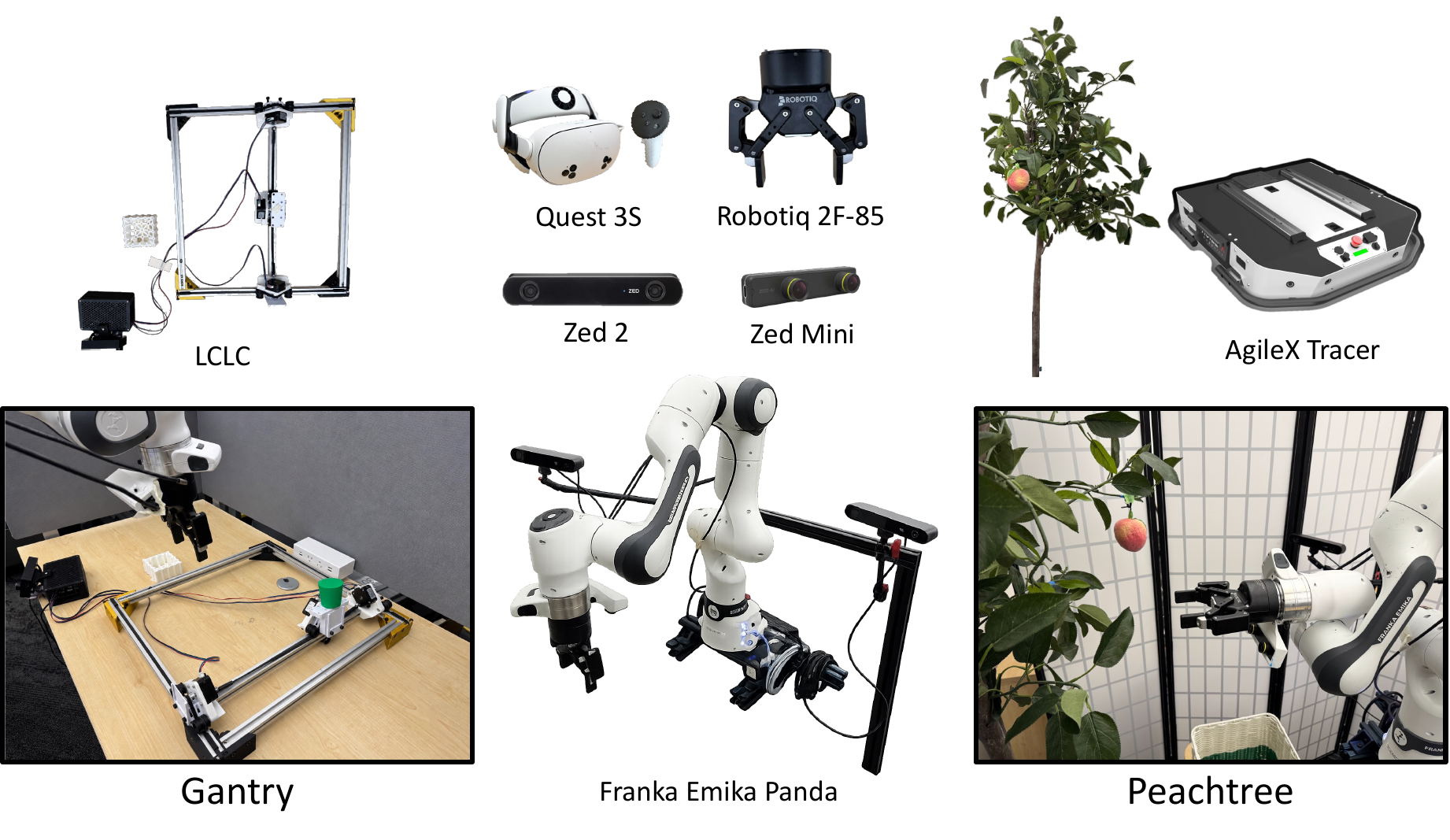}
    \caption{\textbf{SIDO real-world evaluation system.} Both platforms use a Franka arm with the shared perception and gripper stack shown in the center: a ZED~2 for the agent view, a ZED~Mini for the wrist view, and a Robotiq 2F-85 parallel-jaw gripper. \emph{Gantry (left):} a low-cost LCLC gantry translates a capsule, which the arm picks and places into a box. \emph{Peachtree (right):} an AgileX Tracer mobile base sways an artificial peach tree, from which the arm picks a peach and drops it into a basket.}
    \label{fig:SIDO_hardware}
\end{figure}

\paragraph{Manipulator and gripper.} SIDO real-world evaluation system uses a 7-DoF Franka Emika Panda arm under the DROID~\cite{khazatsky2025droidlargescaleinthewildrobot} setup, fitted with a Robotiq 2F-85 parallel-jaw gripper. The policy  is rolled out at $20$\,Hz. Static demonstrations are teleoperated with a Meta Quest~3S headset on top of the DROID setup.

\paragraph{Perception.} A ZED~2 stereo camera on the right supplies the agent view and a wrist-mounted ZED~Mini supplies the wrist view. Both stream RGB images to the policy, downsampled by a factor of $2$ to $640 \times 360$. For object pose estimation, FoundationPose runs on the agent view ZED~2 only.

\paragraph{Gantry.} The object is moved by the Lowest Complexity Laser
Cutter (LCLC)~\cite{cho2020lclc}, an open-source low-cost gantry system. The target is a cylindrical capsule, $5$\,cm tall with a $5$\,cm top and $4$\,cm bottom diameter, and the task places it into a $9 \times 9$\,cm white box. The stage translates the capsule over a $23 \times 43$\,cm workspace at $1.5$\,cm/s. We collect $30$, $22$, and $64$ static demonstrations whose object placements are uniformly distributed over the x-axis, y-axis, and random workspace ranges respectively, the random condition using an $8 \times 8$ grid.

\paragraph{Peachtree.} The object is moved by an AgileX Tracer differential-drive mobile base, which carries a $180$\,cm artificial peach tree and translates it back and forth along a $30$\,cm path at $1.5$\,cm/s, producing a swaying motion of the peach. The target is a $7 \times 7 \times 9$\,cm artificial peach, and the task drops it into a $25 \times 35$\,cm basket below.

\section{Object Motion Patterns}
\label{app:motion}
\paragraph{Simulation.} All five object motion patterns confine the source object to a $25 \times 25$\,cm region in the XY plane, the object moving $2$\,cm/s.
\begin{enumerate}
    \item \textbf{Static.} No motion.
    \item \textbf{X-axis.} Translation along the $x$-axis, reflecting off the workspace bounds.
    \item \textbf{Y-axis.} Translation along the $y$-axis, reflecting off the workspace bounds.
    \item \textbf{Circle.} A circular path of radius $5$\,cm, with an angular step set so the arc length per step matches the $2$\,cm/s speed of the other patterns.
    \item \textbf{Random.} Translation in a direction resampled uniformly every $16$ steps, reflecting off the workspace bounds.
\end{enumerate}
\paragraph{Real-world.} We evaluate on two real-world tasks, the object moving $1.5$\,cm/s at test time. On \textbf{Gantry} the stage translates the capsule under four patterns:
\begin{enumerate}
    \item \textbf{Static.} No motion.
    \item \textbf{X-axis.} Translation along the $x$-axis over the $23$\,cm workspace range, back and forth.
    \item \textbf{Y-axis.} Translation along the $y$-axis over the $43$\,cm workspace range, back and forth.
    \item \textbf{Random.} A new target within the workspace resampled every $31$\,cm of travel, about $20.7$\,s at $1.5$\,cm/s, reflecting off the workspace bounds along the way.
\end{enumerate}
On \textbf{Peachtree} the mobile base moves the tree under two patterns:
\begin{enumerate}
    \item \textbf{Static.} No motion.
    \item \textbf{Dynamic.} The base translates the tree back and forth along its $30$\,cm path, swaying the peach.
\end{enumerate}

\section{Object Pose Predictors}
\label{app:predictors}

At deployment the predictor supplies the object's future position $\hat{P}_{t+T_p}$ into the object-pose channel. Holding the policy fixed, we instantiate three predictors of increasing capacity: a parameter-free \emph{finite-difference} predictor, an \emph{MLP}, and a \emph{flow-equivariant recurrent network (FERNN)}. All read a history of base-frame object positions $P_{t-h}{:}P_t$ obtained by transforming the FoundationPose camera frame pose through the agent view calibration extrinsic, and all output the predicted future pose over the prediction horizon $T_p = 16$. The two learned predictors are trained on five minutes of moving-object trajectories with no policy retraining. We report forecast accuracy as the Final Displacement Error, $\text{FDE} = \lVert \hat{P}_{t+T_p} - P_{t+T_p} \rVert$, averaged over the pre-grasp window ($t, t+T_p \le T_g$).

\paragraph{Finite-difference.}
Finite-difference predictor looks back over the last $3$ observations and, independently for $x$ and $y$, takes the median first difference. The prediction steps along the direction of those median differences with magnitude $v\,T_p$.

\paragraph{MLP.}
The MLP regresses $\widehat{\Delta}$ from the 8-step history, taking the per-step positions and velocities as input through two hidden layers of width $128$. It learns the displacement magnitude from data but carries no motion-structure prior.

\paragraph{FERNN.}
FERNN~\cite{keller2025flowequivariantrecurrentneural} reasons over the same $8$-step history in velocity space. A fixed bank of $K=25$ velocity hypotheses tiles the $x$--$y$ plane; a shared recurrent cell integrates each observed velocity in every hypothesis' co-moving frame, and a softmax gate pools them into the predicted step velocity, scaled by $T_p$. Because each hypothesis sees only its own residual, a shift in object velocity permutes the hypotheses rather than changing the computation, so the estimate re-acquires quickly after a direction change.

\begin{figure}[!ht]
  \centering
  \includegraphics[width=0.8\textwidth]{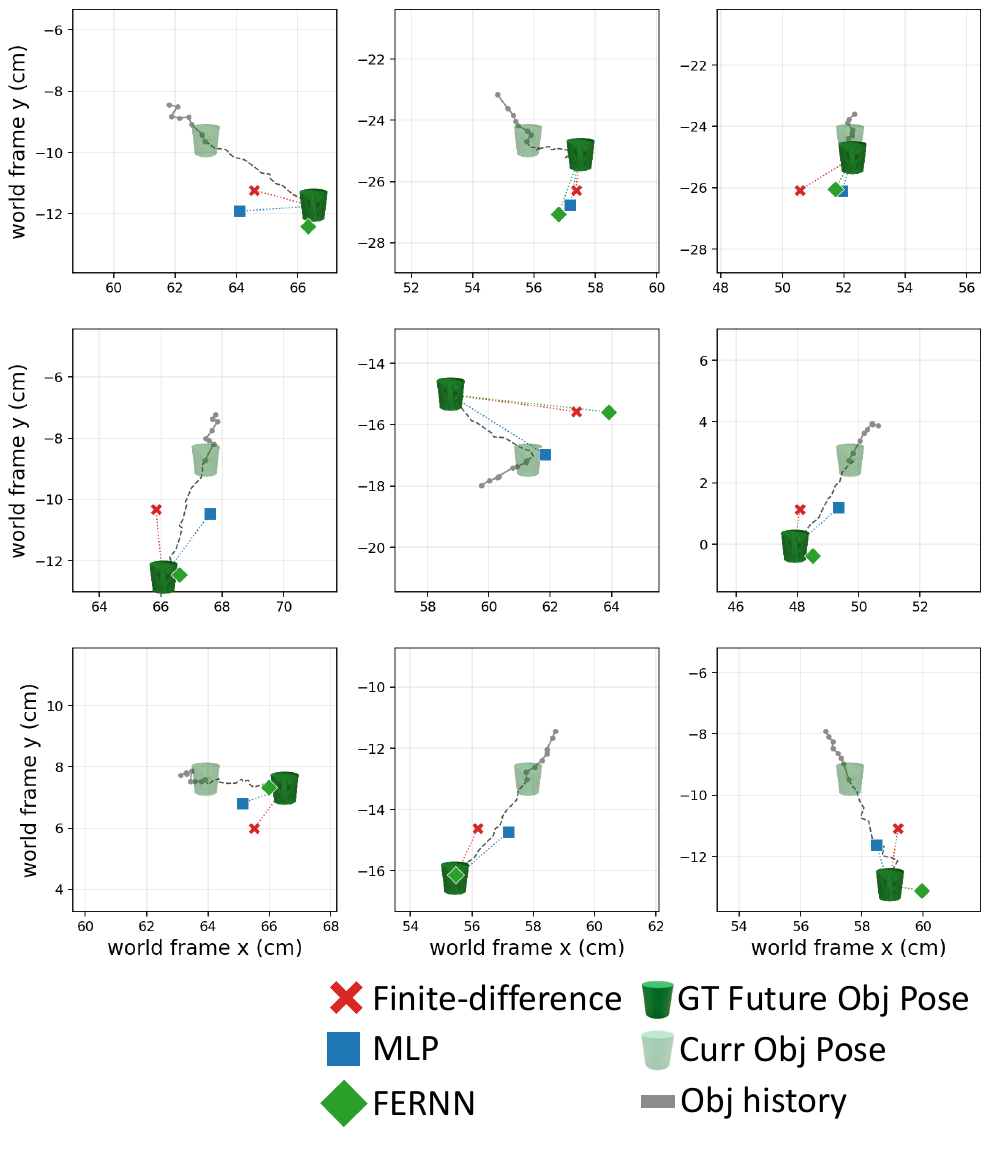}
  \caption{\textbf{Visualization of object pose predictor forecasts on Gantry random.} Each panel shows pre-grasp snapshots in the world frame: the gray dashed line traces the object's recent pose history, the light green marks the current object pose and the dark green the ground-truth future pose one horizon $T_p$ ahead. The three predictors forecast that future pose from the history, finite-difference (red), MLP (blue), and FERNN (green). FERNN lands closest to the ground-truth future pose across panels.}
  \label{fig:predictor_visualization}
\end{figure}

\section{H-SIDO for Relative Actions}
\label{app:heuristic_cf_rel}

Eq.~\eqref{eq:heuristic_cf} shifts each waypoint directly because absolute actions are world-frame positions. Relative actions are step increments, so the same per-step shift would translate every later waypoint with it. We instead apply the ramp in cumulative-path space and index it by arc length, which keeps the morphed trajectory the same shape as the demonstration. The input is a pre-grasp chunk of relative actions $a_{t:t+T_p-1}$, the displacement $\delta$, and the action horizon $T_a$. We write $\epsilon$ for a small numerical constant.

First, integrate the chunk into a path of waypoints, $c_k = \sum_{j<k} a_{t+j}$ with $c_0 = \mathbf{0}$. Then ramp by arc length: with $\ell_k$ the path length up to waypoint $k$, set $\rho_k = \min(\ell_k / \ell_{T_a},\, 1)$, returning the chunk unchanged if $\ell_{T_a} < \epsilon$. Finally, shift each waypoint and difference back,
\begin{equation}
    \tilde{c}_k = c_k + \rho_k\, \delta,
    \qquad
    \tilde{a}_{t+k} = \tilde{c}_{k+1} - \tilde{c}_k.
    \label{eq:app_shift}
\end{equation}
Since $\rho_k = 1$ at both the action horizon terminus and the prediction horizon terminus, the end-effector terminus is displaced by $\delta$, preserving $r_t(T_a)$ and $r_t(T_p)$ as the absolute morph does.

\section{SPARC Smoothness Metric}
\label{app:sparc}

Appendix~\ref{app:hsido_dynasido} measures trajectory smoothness with SPARC 
(spectral arc length)~\cite{balasubramanian2015smoothness}, a refinement of the
earlier SAL measure~\cite{balasubramanian2012robust}. Both work in the frequency
domain: a smooth motion has a compact Fourier magnitude spectrum, so the arc
length of that spectrum is a smoothness proxy, with shorter arcs meaning smoother
motion.

Let $V(\omega)$ be the Fourier magnitude spectrum of a speed profile $v_t$,
normalized by its zero-frequency value,
$\hat{V}(\omega) = V(\omega)/V(0)$, so the measure is invariant to the scale of
the motion. SAL integrates the arc length of $\hat{V}$ up to a cutoff
$\omega_c$,
\begin{equation}
    \text{SAL} \triangleq
    -\int_{0}^{\omega_c}
    \left[
        \left(\tfrac{1}{\omega_c}\right)^2
        + \left(\frac{d\hat{V}(\omega)}{d\omega}\right)^2
    \right]^{1/2} d\omega ,
\end{equation}
where the $1/\omega_c$ term normalizes for the cutoff and the minus sign makes
smoother trajectories score higher. SPARC sets $\omega_c$ adaptively, as the
lowest frequency beyond which $\hat{V}$ stays under a threshold $\bar{V}$, capped
at $\omega_c^{\max}$,
\begin{equation}
    \omega_c \triangleq
    \min \Bigl\{
        \omega_c^{\max},\;
        \min \bigl\{
            \omega :
            \hat{V}(r) < \bar{V}
            \;\;\forall\, r > \omega
        \bigr\}
    \Bigr\} ,
\end{equation}
which keeps high-frequency noise from inflating the arc length. We zero-pad each
speed trajectory by a factor of $K=4$ for frequency resolution, use
$\omega_c^{\max}=20$ and $\bar{V}=0.05$, and compute the arc length as the summed
Euclidean distance between successive spectrum points. Higher SPARC means
smoother end-effector motion.

\section{Implementation Details}
\label{app:implementation}

\begin{table}[ht]
\centering
\small
\setlength{\tabcolsep}{4pt}
\renewcommand{\arraystretch}{1.05}
\caption{SIDO implementation settings.}
\label{tab:sido_settings}
\begin{tabular}{@{}l l@{}}
\toprule
Setting & Value \\
\midrule
Horizons
& $T_p=16,\;T_a=8$ \\
Augmentation ratio
& $\alpha=0.2$ static, $0.8$ counterfactual \\
Displacement scale
& $v=0.001$ simulation units/step,
  $\|\delta\|=0.016$ simulation units \\
Dynamics model
& Three-layer MLP \\
MPPI
& $128$ samples, $10$ iterations, $\beta=0.5$ \\
\bottomrule
\end{tabular}
\end{table}

\end{document}